\PassOptionsToPackage{table}{xcolor}
 \documentclass[sigconf]{acmart}
\usepackage{CJK}
\usepackage{booktabs}
\usepackage{multirow}
\usepackage{amsmath}

\usepackage{amssymb}
\usepackage{algorithm}
\usepackage{algorithmic}
\usepackage{makecell}
\usepackage{graphicx}
\usepackage{subcaption}
\usepackage[table]{xcolor}
\usepackage{enumitem}
\AtBeginDocument{%
  }

\setcopyright{acmlicensed}
\copyrightyear{2026}
\acmYear{2026}
\setcopyright{cc}
\setcctype{by}
\acmConference[ICMR '26]{International Conference on Multimedia Retrieval}{June 16--19, 2026}{Amsterdam, Netherlands}
\acmBooktitle{International Conference on Multimedia Retrieval (ICMR '26), June 16--19, 2026, Amsterdam, Netherlands}
\acmDOI{10.1145/3805622.3810601}
\acmISBN{979-8-4007-2617-0/2026/06}

\begin{document}

\author{Jiale Huang}
\affiliation{%
 \institution{Shandong University}
 \city{Jinan}
 \state{Shandong}
 \country{China}}
 \email{huangjiale359@mail.sdu.edu.cn}

\author{Zixu Li}
\affiliation{%
 \institution{Shandong University}
 \city{Jinan}
 \state{Shandong}
 \country{China}}
 \email{lizixu.cs@gmail.com}

\author{Zhiwei Chen}
\affiliation{%
 \institution{Shandong University}
 \city{Jinan}
 \state{Shandong}
 \country{China}}
 \email{zivczw@gmail.com}
 
\author{Zhiheng Fu}
\affiliation{%
 \institution{Shandong University}
 \city{Jinan}
 \state{Shandong}
 \country{China}}
 \email{fuzhiheng8@gmail.com}

\author{Chunxiao Wang}
\affiliation{%
 \institution{Qilu University of Technology (Shandong Academy of Sciences)}
 \city{Jinan}
 \state{Shandong}
 \country{China}}
 \email{wangchx@sdas.org}
 
\author{Yupeng Hu}
\authornote{Yupeng Hu is the corresponding author.}
\affiliation{%
 \institution{Shandong University}
 \city{Jinan}
 \state{Shandong}
 \country{China}}
 \email{huyupeng@sdu.edu.cn}

\renewcommand{\shortauthors}{Huang et al.}
\def\modelname{\mbox{IMAGINE} }

\begin{abstract}
Composed Video Retrieval (CVR) is designed to retrieve a target video that matches a reference video modified by a modification text. While existing methods explore cross-modal correspondences, they often assume modified objects appear directly in videos. However, modification texts frequently describe concepts not explicitly presented but implicitly expressed through semantically related visual cues (e.g., ``cake'' implying ``birthday party''). Current approaches typically rely on aligning explicit feature representations within the concrete space, neglecting critical latent associations. To address this, we propose an adapt\textbf{I}ve sche\textbf{M}a-Im\textbf{AG}ery enhanced compos\textbf{I}tional \textbf{NE}twork (\textbf{IMAGINE}). Unlike standard explicit matching, IMAGINE materializes implicit semantics (termed schema imagery) via dynamic multimodal prototypes. These prototypes capture shared latent concepts to adaptively modulate visual features, effectively injecting implicit guidance into the retrieval process. By bridging the gap between explicit visual contents and implicit retrieval intentions, IMAGINE achieves state-of-the-art performance in both CVR and Composed Image Retrieval (CIR) across three widely used benchmarks.
\end{abstract}



\title{IMAGINE: Adaptive Schema-Imagery Enhanced Composition for Composed Video Retrieval}

\maketitle

\section{Introduction}
\begin{figure}[ht]
  \begin{center}
\includegraphics[width=\linewidth]{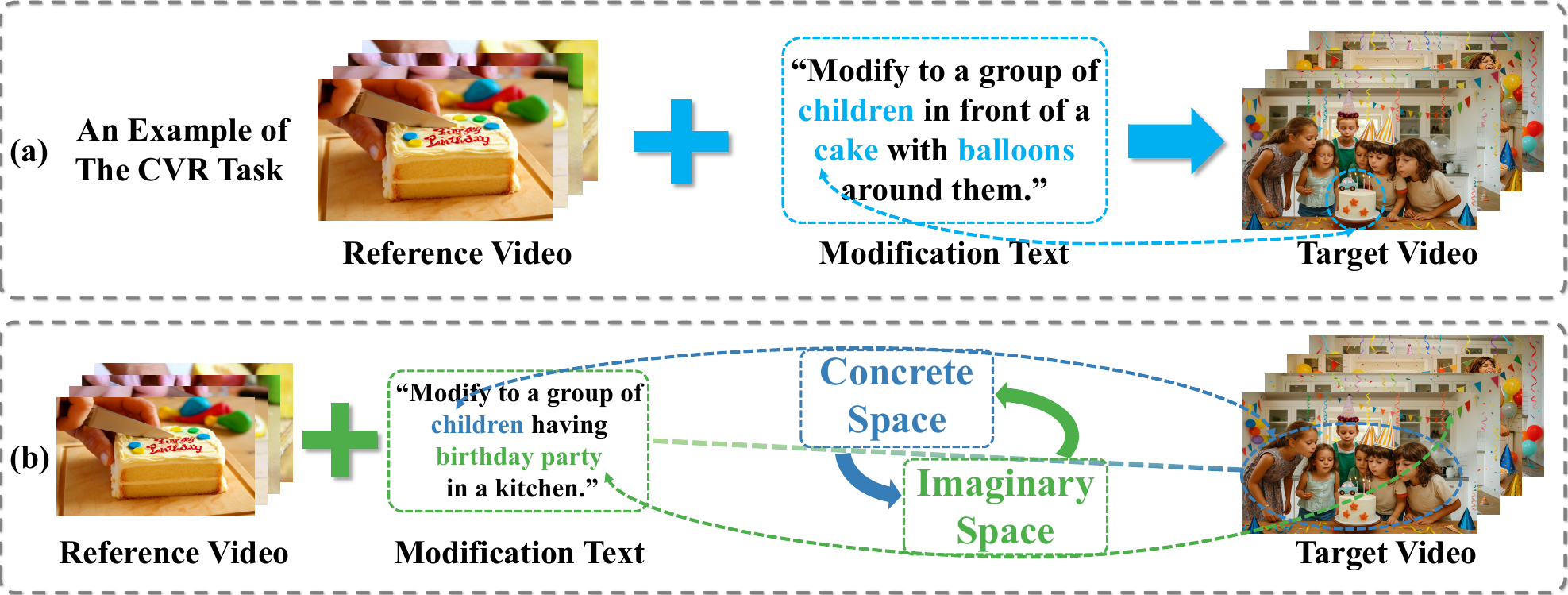}
  \end{center}
  \caption{Example of CVR task and IMAGINE's main idea.}
  \label{fig:intro}
\end{figure}

The CVR task aims to retrieve target videos from a large video library based on a multimodal query consisting of a reference video and a modification text, which aligns with the semantic intent of the query. Fig.\ref{fig:intro}(a) presents an example of CVR. Unlike traditional uni-modal or cross modal retrieval~\cite{jiang2024prior,STABLE,R3}, the text in CVR expresses the modification requirements for the reference video, rather than providing a complete description. This allows the CVR task to more naturally express complex retrieval intentions~\cite{HABIT,OFFSET,cirr,TME,HINT}. In scenarios such as efficient system~\cite{lu2024robust,proxmo,huang2025enhancing,zeng2025bridging,chen2026grace,YANG2026114166,bi2026the,jiang2026drpdistilledreasoningpruning,huangnodes,li2024synergized,jiang2025mined,li2025domain,ERASE} and computer vision~\cite{yang2026muse,lu2024mace,zhang2026adaptive,wan2025hyperion,jiang2026beyond,zhang2026finsentllm,li2024incomplete,li2026modality,curriculum_rlaif,li2025cross,tian2025core,li2025set}, CVR demonstrates significant application value.

Recent CVR research~\cite{covr, covr_enrich, REFINE,covr-2,ReTrack,fdca} focuses on capturing complementary information via cross-modal interaction but relies heavily on Explicit Object Correspondence. We analyze this limitation by formalizing retrieval into two dimensions: \textbf{1) concrete space}, consisting of explicitly presented physical objects such as cakes; and \textbf{2) imagery space}, comprising latent semantic associations and scene atmospheres evoked by these objects (e.g., associating a cake with a birthday party). Mainstream methods remain confined to the concrete space, assuming object-level correspondence for the modification text in the reference/target video~\cite{covr,covr-2}. 
While recent work like HUD~\cite{hud} attempts to mitigate retrieval ambiguity via uncertainty modeling, it remains limited to aligning explicit visual cues rather than constructing the latent imagery space.
However, real-world scenarios frequently exhibit semantic-visual misalignment~\cite{ke2025early,limn,jiang2025kore,jiao2026large,yu2025vismem,jiang2026scribe,bi2026echorl,huang2023robust,zhang2023spot,shi2026multiscenario,maspo,zhang2026multivariate,huangrevisiting,yu2025visual}. As shown in Fig.~\ref{fig:intro}(b), while the instruction aims to construct a birthday party imagery, no direct entity exists in the reference video or target video; the scene is inferred solely from cues like ribbons and cakes. While humans naturally bridge these spaces, existing models lack explicit modeling of the imagery space, failing to establish the inference path from concrete entities to imagery semantics. This lack of Cross-space Reasoning neglects critical latent semantic cues, significantly limiting retrieval accuracy in complex scenarios~\cite{FineCIR}.

To bridge the gap between concrete perception and imagery reasoning, two core challenges must be addressed. \textbf{C1: How to construct a scalable semantic memory to materialize the imagery space?} As an abstract concept, the imagery space requires materialization into computable representations, yet existing methods often lack mechanisms to extract such auxiliary semantics. We propose that the key lies in designing a dynamic memory structure to parameterize the abstract imagery space, enabling the model to actively retrieve related latent semantics when visual cues are absent. \textbf{C2: How to achieve adaptive cross-space fusion during composition?} While introducing the imagery space enriches semantics, it introduces potential noise risks. The importance of concrete versus imagery features varies significantly across queries, such as changing color versus altering scene atmosphere. Therefore, simple feature concatenation is suboptimal. A confidence-based Dual-space Modulation mechanism is required to adaptively control the intervention of imagery information, enhancing semantics while minimizing interference with original visual features.

To address this, we proposes the adapt\textbf{I}ve sche\textbf{M}a-im\textbf{AG}ery enhanced compos\textbf{I}tional \textbf{NE}twork (\textbf{IMAGINE}). The core idea is to capture and utilize semantically related imagery space to assist in modeling the original composed features, while simultaneously guiding feature alignment in the concrete space to achieve better retrieval performance. The framework consists of three core modules: \textit{(a) Schema Imagery Construction}, which adaptively updates the inter-modality prototype library through an attention mechanism to generate imagery vectors, complementing the representations in the concrete space; \textit{(b) Imagery-guided Multimodal Composition}, which comprehensively considers both the concrete space and imagery space during the composition process and uses imagery space to modulate the composed features of the concrete space; \textit{(c) Dual Space Alignment}, which leverages the complementarity between the concrete space and imagery space to perform dual-channel alignment between the composed features and the target video. Furthermore, extensive experimental results on three widely used benchmark datasets, covering both CVR and CIR tasks, demonstrate the superiority of our IMAGINE.

Our contributions are summarized as follows:
\begin{itemize}[leftmargin=8pt]
    \item We propose IMAGINE, which explicitly models the imagery space to capture latent associations, effectively addressing the semantic-visual misalignment issue.
    \item We design \textit{Schema Imagery Construction} to build dynamic multimodal prototypes, and employ a confidence-based modulation mechanism to adaptively inject implicit semantic guidance into the cross-modal composition process.
    \item IMAGINE achieves state-of-the-art performance on three widely used benchmarks, demonstrating superior generalization capabilities across both CVR and CIR tasks.
\end{itemize}

\begin{figure*}[ht]
  \begin{center}
  \includegraphics[width=0.95\linewidth]{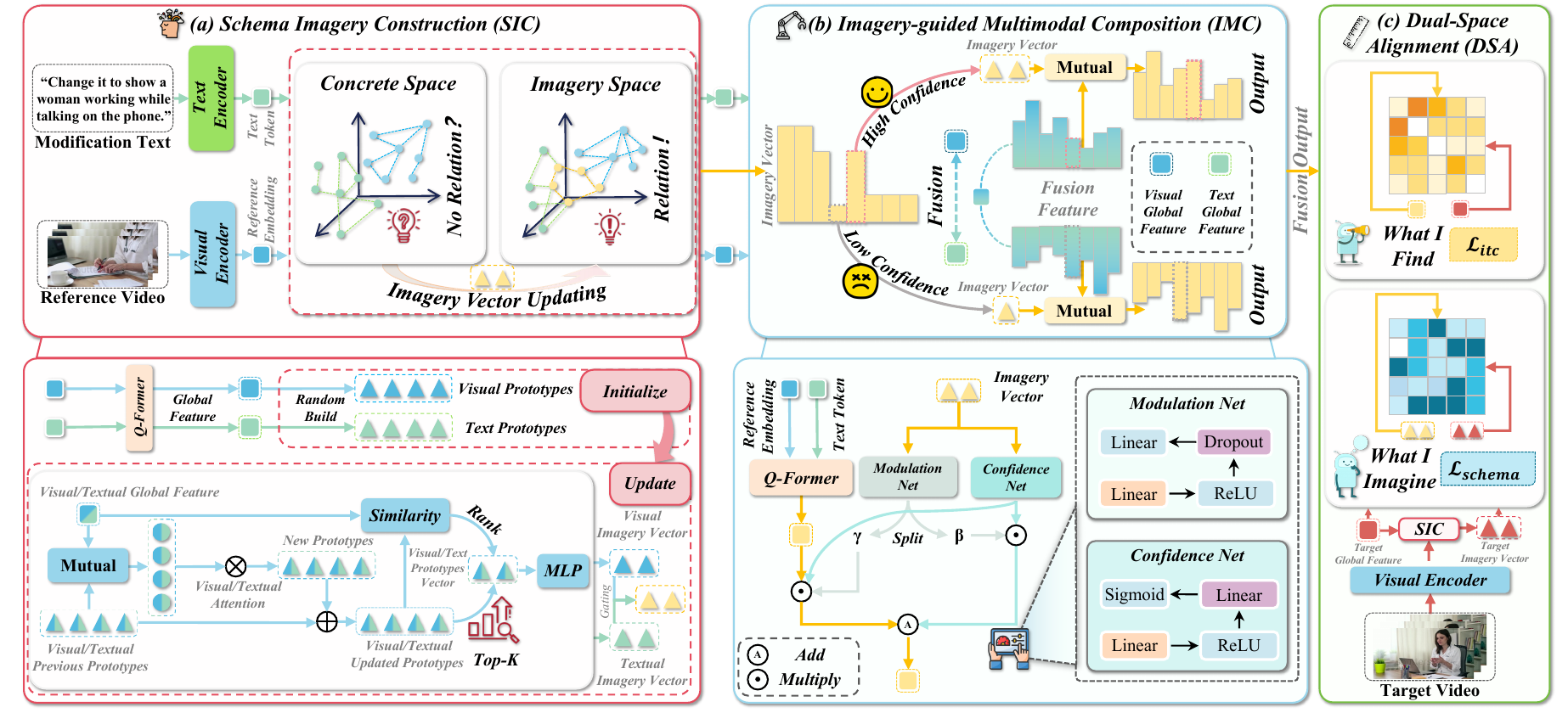}
  \end{center}
  \caption{IMAGINE Framework: \textbf{(a)} Schema Imagery Construction, \textbf{(b)} Imagery-guided Multimodal Composition, \textbf{(c)} Dual Space Alignment.}
  \label{fig:IMAGINE}
\end{figure*}

\section{Related Work}

\textbf{CIR and CVR task.} Composed Image Retrieval (CIR) retrieves target images based on multimodal queries. Existing methods generally fall into two categories~\cite{INTENT,Air-Know,ConeSep}. The first utilizes standard encoders (e.g., ResNet, BERT) coupled with specific cross-modal alignment modules. Specifically, Zhang \textit{et al.}~\cite{JAMMA} utilized graph attention networks to model interaction relationships, Yang \textit{et al.}~\cite{JPM} proposed a joint prediction module to align visual-textual discrepancies, and Zhang \textit{et al.}~\cite{eer} adopted an erase-and-supplement strategy to achieve semantic fusion. The second category leverages Vision-Language Pre-training (VLP) models such as CLIP and BLIP~\cite{yu2025vismem,COMBINER,TEMA}. For instance, Zhao \textit{et al.}~\cite{Prog-Lrn} proposed an adaptive weighting strategy for dynamic query composition, while Zhu \textit{et al.}~\cite{amc} introduced a multi-expert collaborative network.
Similar to CIR, Composed Video Retrieval (CVR) utilizes multimodal queries but distinguishes itself by replacing images with videos. Ventura \textit{et al.}~\cite{covr, covr-2} leveraged the generalized knowledge of BLIP and BLIP-2 to comprehend video-text pairs for effective adaptation. Thawakar \textit{et al.}~\cite{covr_enrich} facilitated multimodal understanding by generating detailed video captions to mitigate semantic ambiguity. However, these methods predominantly rely on explicit semantic correspondence, assuming that retrieval cues are directly visible within the reference video. This overlooks the necessity for implicit semantics in scenarios where the target must be inferred conceptually. To address this, we propose ``Schema Imagery'' to mine latent associations and enhance retrieval precision.

\noindent \textbf{Generative Imagination Enhancement in Cross-modal Reasoning.}
Generative imagery enhancement improves semantic consistency in cross-modal reasoning by constructing computable imagery representations to complement missing visual information~\cite{jiang2026cornerstones,sun2024robust,EgoAction,TempRet,xu2026learningusetoolsjust,zhao2026luve}. This field has evolved from basic prototype synthesis to active logical prediction. Early research focused on synthesizing visual prototypes using creativity-inspired methods or inductive learning, simulating human associative processes by establishing correlations between attributes and features~\cite{zhang2026optimalteacherpersonalizeddata,li2023cross,xie2025chat,cao2026task,li2025generation,huang2025final,se_agent,sun2023hierarchical,EgoAdapt}. 
With the rise of large-scale pre-trained models, the research focus shifted toward utilizing divergent thinking to materialize implicit semantics \cite{Ldre}, thereby addressing perceptual biases caused by missing visual information.
Recent trends have further introduced World Models and multi-agent collaboration mechanisms \cite{gao2024eraseanything,shi2026mmerror,OmniEgo-R,lu2026chordedit,ren2025all,lu2022copy}, aiming to achieve more prospective active information prediction and native visual reasoning. 
Despite these advancements, the potential of adaptive imagery construction in composed retrieval remains to be fully explored~\cite{qin2023cross,yuan2025prototype}. In this work, we introduce an imagery-inspired enhancement strategy for CVR tasks, encouraging the model to perform cross-space semantic alignment and reasoning during feature composition.

\section{Methodology}\label{sec:method}

IMAGINE's key innovation is leveraging semantically related imagery space to assist in modeling composed features and guide feature alignment in the concrete space for better retrieval performance. As shown in Fig.~\ref{fig:IMAGINE}, IMAGINE includes three main modules:
\textit{(a) Schema Imagery Construction (SIC)}: Builds an inter-modality imagery prototype library and generates stable imagery vectors to improve concrete representation.
\textit{(b) Imagery-guided Multimodal Composition (IMC)}: Modulates composed features during composition with imagery space guidance, adding semantic information.
\textit{(c) Dual Space Alignment (DSA)}: 
Optimizes retrieval accuracy via dual-space alignment of composed features and target video.

\subsection{Problem Formulation}

The CVR task aims to retrieve target videos that match the query conditions based on a multimodal query consisting of a reference video and modification text. Specifically, given a triplet $\mathcal{T}=\left\{\left(x_{r},t_{m},x_{t}\right)_{i}\right\}_{i=1}^{M}$, where $x_{r},t_{m},x_{t}$ represent the reference video, modification text, and target video, respectively, our goal is to optimize metric learning such that the multimodal query ($x_{r},t_{m}$) is as close as possible to the corresponding target video $x_{t}$. This can be expressed in a formula as $\mathcal{F}\left(x_r,t_m\right){\rightarrow}\mathcal{F}\left(x_t\right)$.

\subsection{Schema Imagery Construction}

\noindent As shown in Fig.~\ref{fig:IMAGINE}(a), the \textit{Schema Imagery Construction} module constructs a stable and accurate imagery space, providing guidance for cross-modal composition. It includes two key steps: the construction of an inter-modality imagery prototype library and the generation of shared imagery vectors across modalities.

\noindent \textit{\textbf{Step~1: Construction of Inter-Modality Imagery Prototype Library.} }
Due to the heterogeneity between text and video, building a cross-modal imagery space is challenging. We address this by constructing an imagery prototype library for each modality and iteratively updating it during training to refine imagery vectors. 

We begin by building the video imagery prototype library, using the reference video to describe the process, as it shares a common prototype library with the target video. 
To leverage the imagery space's generalization, we extract global features, $\textbf{F}_r \in \mathbb{R}^{D}$, from $N_f$ randomly sampled frames of the reference video $x_r$ using BLIP-2~\cite{blip-2} with average pooling. Initially, when $k=0$, the model has not yet learned the reference video's semantics; thus, we initialize a global imagery prototype library $P_r \in \mathbb{R}^{N \times D}$ shared across the dataset. We set $N=32$ to align with the Q-Former's query tokens, enabling the prototypes to inherit the pre-trained semantic diversity of these slots without additional constraints.

For iterations $k>1$, we employ a batch-wise momentum update strategy to maintain diversity. Given a batch of $B$ reference videos $\{\textbf{F}_r^{(b)}\}_{b=1}^B$, we calculate the attention weight $\textbf{a}_{b,i}$ between the $b$-th video and the $i$-th prototype $\textbf{P}_{r,i}^{k-1}$,

\begin{equation}
\fontsize{9pt}{9pt}
    \textbf{a}_{b,i} = \frac{\exp(\text{s}(\textbf{F}_r^{(b)}, \textbf{P}_{r,i}^{k-1}) / \tau)}{\sum_{j=1}^{B} \exp(\text{s}(\textbf{F}_r^{(j)}, \textbf{P}_{r,i}^{k-1}) / \tau)},
\end{equation}
where $\textbf{a} \in \mathbb{R}^{N}$, and $\tau$ denotes the temperature coefficient.

Subsequently, each prototype aggregates relevant semantics from the batch samples,
\begin{equation}
    \textbf{P}_{r,i}^{k} = \rho \textbf{P}_{r,i}^{k-1} + (1-\rho) \frac{\sum_{b=1}^{B} \textbf{a}_{b,i} \cdot \textbf{F}_{r}^{(b)}}{\sum_{b=1}^{B} \textbf{a}_{b,i}},
\label{rhomod}
\end{equation}
where $\rho \in [0,1]$ is the modulation coefficient.Unlike single-instance updates, this batch-wise aggregation enables prototypes to attend to distinct sample subsets, preserving semantic diversity.

Similarly, for the text modality, after $k$ iterations, we can construct the text imagery prototype library $\mathbf{P}_{m}^{k} \in \mathbb{R}^{N \times D}$.

\noindent \textit{\textbf{Step 2: Generation of shared imagery vectors across modalities.}}
We obtained the imagery prototype libraries for both video and text to generate inter-modality imagery vectors to represent shared semantics by calculating similarity scores between each modality's concrete semantics and its corresponding prototype library,

\begin{equation}
    \fontsize{9pt}{9pt}
    s_{\text{ref,n}} = \frac{\textbf{F}_{r} \cdot \textbf{P}_{r,n}^k}{\|\textbf{F}_r\| \|\textbf{P}_{r,n}^k\|}, \quad
    s_{\text{mod,n}} = \frac{\textbf{F}_m \cdot \textbf{P}_{m,n}^k}{\|\textbf{F}_m\| \|\textbf{P}_{m,n}^k\|},
\label{}
\end{equation}
where $n \in \{1, \dots, N\}$, $\mathbf{P}_{r, n}^k$ and $\mathbf{P}_{m,n}^k$ denotes the $n$-th prototype vector in $\mathbf{P}_r^k$ and $\mathbf{P}_m^k$.

Based on scores, we select the top-$T$ imagery prototypes most relevant to the concrete semantics of both visual and text modalities. Let $\mathcal{J}_{ref}$, $\mathcal{J}_{mod}$ be the set of indices corresponding to the top-$T$ values in the score vector $s_{\text{ref}}$ and $s_{\text{mod}}$,
\begin{equation}
    \fontsize{9pt}{9pt}
    c_{\text{ref}} = \{ \mathbf{P}_{r, j}^k \mid j \in \mathcal{J}_{\text{ref}} \}, \quad c_{\text{mod}} = \{ \mathbf{P}_{m, j}^k \mid j \in \mathcal{J}_{\text{mod}} \},
\label{}
\end{equation}
where $c_{\text{ref}}$ and $c_{\text{mod}}$ include the top $T$ video and text imagery prototypes, respectively. These sets are stacked and pooled into imagery vectors representing the inter-modality imagery space, as below,
\begin{equation}
    \fontsize{9pt}{9pt}
    \textbf{m}_r = \frac{1}{T} \sum_{\textbf{P}_r^k \in c_{\text{ref}}} \textbf{P}_r^k,\quad
    \textbf{m}_m = \frac{1}{T} \sum_{\textbf{P}_m^k \in c_{\text{mod}}} \textbf{P}_m^k, 
\label{}
\end{equation}
where $\textbf{m}_{r}$ and $\textbf{m}_{m} \in \mathbb{R}^{D}$ represent the imagery vectors of the reference video and modification text, respectively. Similarly, we can obtain the imagery vector $\textbf{m}_{t}$ for the target video.

Finally, we fuse the imagery vectors of the video and text to obtain a unified imagery representation shared between the modalities, providing the semantic foundation for the subsequent compositional process, as formulated below,
\begin{gather}
    \fontsize{9pt}{9pt}
    w_r, w_m = \text{MLP}(\textbf{m}_r, \textbf{m}_m),
    \textbf{m} = w_r \cdot \textbf{m}_r + w_m \cdot \textbf{m}_m,
\label{}
\end{gather}
where $w_r$ and $w_m$ are the fusion weights of the reference video and modification text, respectively.

\subsection{Imagery-guided Multimodal Composition} 

\noindent To improve composed feature representation, we introduce \textit{Imagery-guided Multimodal Composition (IMC)}, which uses the shared inter-modality imagery vector to guide concrete space features, leveraging the complementary strengths of both spaces for better semantic fusion.
Using the Q-Former in BLIP-2, we fuse the reference video $x_r$ and modification text $t_m$ to generate the original concrete composed feature $\textbf{F}_c^{ori} \in \mathbb{R}^{D}$, as shown below,
\begin{equation}
    \fontsize{9pt}{9pt}
    \textbf{F}_c^{ori}=\operatorname{Avg}(\operatorname{Q-Former}(\varPhi_\mathbb{I}(x_r),\varPhi_\mathbb{T}(t_m))),
\label{}
\end{equation}
where $\operatorname{Avg}(\cdot)$ represents average pooling, and $\varPhi_{\mathbb{I}}$ and $\varPhi_{\mathbb{T}}$ denote the visual encoder and text tokenizer, respectively.

To adjust key dimensions of the concrete composed feature using imagery semantics, we input the shared inter-modality imagery vector $\textbf{m}$ into a modulation network (detailed in Fig.~\ref{fig:IMAGINE}(b)), which generates parameters for feature modulation, formulated as,

\begin{equation}
    \fontsize{9pt}{9pt}
    [\gamma_{\text{raw}}, \beta_{\text{raw}}] = \text{Modulation-Net}(\textbf{m}),
\label{}
\end{equation}
where $\gamma_{raw},\beta_{raw}\in\mathbb{R}^{D}$ are the modulation direction and shift coefficients, corresponding to the feature dimension $D$. 
To prevent low-correlation imagery from affecting the concrete feature, we use a confidence network to assess the credibility of the scaled modulation. The network implementation is shown in Fig.~\ref{fig:IMAGINE}(b). The process is as follows,
\begin{equation}
    \fontsize{9pt}{9pt}
    \textbf{c} = \text{Sigmoid}(\text{Confidence-Net}(\textbf{m})),
\label{}
\end{equation}
where $\textbf{c} \in \mathbb{R}^{1}$ represents the credibility of the scaled modulation intensity. This credibility is then used to weight the direction coefficient $\gamma_{\text{raw}}$ and shift coefficient $\beta_{\text{raw}}$ as,
\begin{equation}
    \fontsize{9pt}{9pt}
    \gamma = \gamma_{\text{raw}} \odot \textbf{c}, \quad
    \beta = \beta_{\text{raw}} \odot \textbf{c},
\label{}
\end{equation}
where $\odot$ denotes element-wise multiplication with batch broadcasting. Finally, we modulate the concrete composed feature as follows, obtaining the final composed feature, as formulated below,
\begin{gather}
    \fontsize{9pt}{9pt}
    \begin{aligned}
    \textbf{F}_c = \textbf{F}_c^{ori} \odot (1 + \gamma) + \beta.
    \end{aligned}
\label{}
\end{gather}

\subsection{Dual Space Alignment} 

\noindent With the first two modules, we refine the multimodal composed feature representations in both the concrete and imagery spaces. To exploit the complementary benefits of both spaces and align the composed feature with the target video feature, we design \textit{Dual Space Alignment (DSA)}, as shown in Fig.~\ref{fig:IMAGINE}(c).

For the concrete space, we obtain the target video's global feature $\textbf{F}_t \in \mathbb{R}^{D}$ using BLIP-2 and apply a batch-based classification loss to align the multimodal composition with the target feature as,
    \begin{equation}
    \fontsize{9pt}{9pt}
    \mathcal{L}_{itc} = - \frac{1}{B} \sum_{i=1}^{B} 
    \log \frac{\exp \big(s(\textbf{F}_c^i, \textbf{F}_t^i)/\tau \big)}
    {\sum_{j=1}^{B} \exp \big(s(\textbf{F}_c^i, \textbf{F}_t^j)/\tau \big)},
    \label{}
    \end{equation}
where $\textbf{F}_c^i$ is the $i$-th composed global feature, and $\textbf{F}_t^i$ and $\textbf{F}_t^j$ represent the true and candidate target global features, respectively, with $s(\cdot)$ denoting cosine similarity. 
For the imagery space, we enforce semantic consistency between the inferred query imagery $\textbf{m}$ and the ground-truth target imagery $m_t$. Specifically, $\textbf{m}_t$ is obtained by the same shared SIC process as the reference video, ensuring alignment in the same semantic manifold. The schema alignment loss is defined as,
    \begin{equation}
    \fontsize{9pt}{9pt}
    \mathcal{L}_{schema} = \frac{1}{B}\sum_{b=1}^{B} \left( 1 - \cos(\textbf{m}^{}, \textbf{m}_{t}^{}) \right),
    \label{}
    \end{equation}
where $\cos(\cdot, \cdot)$ denotes cosine similarity.

Finally, we obtain the final loss function, formulated as,
    \begin{equation}
    \fontsize{9pt}{9pt}
    \mathbf{\Theta^{*}}=
    \underset{\mathbf{\Theta}}{\arg \min } \left( \mathcal{L}_{itc} + \kappa \mathcal{L}_{\text{schema}}\right),
    \label{optimization}
    \end{equation}
where $\mathbf{\Theta}$ is the IMAGINE parameter to be learned and $\kappa$ are the trade-off hyper-parameters.

\begin{table}[ht]
  \centering
  \small
  \tabcolsep=8pt
  \caption{Performance comparison on the test set of the CVR dataset, WebVid-CoVR, relative to R@$k$($\%$). The overall best results are in bold, while the best results over baselines are underlined.}
    \resizebox{\linewidth}{!}{
    \begin{tabular}{l|cccc|c}
    \hline
    \hline
    \multicolumn{1}{c|}{\multirow{3}{*}{Method}} & \multicolumn{5}{c}{WebVid-CoVR-Test} \\
\cline{2-6}          & \multicolumn{4}{c|}{R@$k$}      & \multirow{2}{*}{Avg.}  \\
\cline{2-5}          & $k$=$1$   & $k$=$5$   & $k$=$10$  & $k$=$50$  &   \\
    \hline
    \multicolumn{6}{c}{\textbf{\textit{\underline{Pre-trianed Models}}}}\\
    CLIP~\cite{clip}\textcolor{gray}  {\scriptsize{ (ICML'21)}} & 44.37 & 69.13 & 77.62 & 93.00 & 71.03  \\
    BLIP~\cite{blip}\textcolor{gray}{\scriptsize{ (ICML'22)}} & 45.46 & 70.46 & 79.54 & 93.27 & 72.18  \\
    
    \hline
    \multicolumn{6}{c}{\textbf{\underline{\textit{CVR Models}}}}\\
    CoVR~\cite{covr}\textcolor{gray}{\scriptsize{ (AAAI'24)}} & 53.13 & 79.93 & 86.85 & 97.69 & 79.40  \\
    CoVR-2~\cite{covr-2}\textcolor{gray}{\scriptsize{ (TPAMI'24)}} & 59.82 & 83.84 & \underline{91.28} & 98.24 & 83.30 \\
    CoVR\_Enrich~\cite{covr_enrich}\textcolor{gray}{\scriptsize{ (CVPR'24)}} &\underline{60.12} & \underline{84.32} & 91.27 & \underline{98.72} & \underline{83.61}  \\
    FDCA~\cite{fdca}\textcolor{gray}{\scriptsize{ (ICLR'25)}} & 54.80 & 82.27 & 89.84 & 97.70 & 81.15  \\
    \hline
       \rowcolor[rgb]{0.949, 0.949, 1.0}
    \textbf{IMAGINE (Ours)} & \textbf{63.51} & \textbf{87.26} & \textbf{92.72} & \textbf{99.03} & \textbf{85.63} \\
    \hline
    \hline
    \end{tabular}%
    }
  \label{tab:cvr}%
\end{table}%

\begin{table*}[ht]
\centering
\caption{Performance comparison on FashionIQ and CIRR relative to R@$k$(\%). The overall best results are in bold, while the second-best results are underlined. The Avg metric in CIRR denotes the mean of all recall metrics.} 
\resizebox{0.95\linewidth}{!}{
\begin{tabular}{l|cc|cc|cc|cc|cccc|ccc|c}
    \hline
    \hline
    \multicolumn{1}{c|}{\multirow{3}{*}{Method}} &  \multicolumn{8}{c|}{FashionIQ}                              & \multicolumn{8}{c}{CIRR} \\
\cline{2-17}                  & \multicolumn{2}{c|}{Dresses} & \multicolumn{2}{c|}{Shirts} & \multicolumn{2}{c|}{Tops\&Tees} & \multicolumn{2}{c|}{Avg} & \multicolumn{4}{c|}{R@$k$} & \multicolumn{3}{c|}{R$_{subset}$@$k$} & \multirow{2}{*}{Avg} \\
\cline{2-16}                  & R@$10$  & R@$50$  & R@$10$  & R@$50$  & R@$10$  & R@$50$  & R@$10$ & R@$50$  & $k$=$1$   & $k$=$5$   & $k$=$10$ & $k$=$50$ & $k$=$1$   & $k$=$2$  & $k$=$3$ &  \\
    \hline
    \hline
    \multicolumn{16}{c}{\textbf{\textit{\underline{CIR Models}}}}\\
    TG-CIR~\cite{tgcir}\textcolor{gray}{\scriptsize{ (ACM MM'23)}}&  45.22 & 69.66 & 52.60 & 72.52 & 56.14 & 77.10 & 51.32 & 73.09 & 45.25 & 78.29 & 87.16 & 97.30 & 72.84 & 89.25 & 95.13 & 80.75 \\
    SADN~\cite{sadn}\textcolor{gray}{\scriptsize{ (ACM MM'24)}} & 40.01  & 65.10  & 43.67  & 66.05  & 48.04  & 70.93  & 43.91 & 67.36 & 44.27  & 78.10  & 87.71 & 97.89 & 72.34 & 88.70  & 95.23  & 80.61\\
    SPRC~\cite{sprc}\textcolor{gray}{\scriptsize{ (ICLR'24)}}  & 49.18 & 72.43 & 55.64 & 73.89 & 59.35 & 78.58 & 54.72 & 74.97 & 51.96 & 82.12 & 89.74 & 97.69 & \textbf{80.65} & \underline{92.31} & \underline{96.60} & \underline{84.44} \\
    LIMN~\cite{limn}\textcolor{gray}{\scriptsize{ (TPAMI'24)}} & 50.72  & 74.52  & 56.08  & 77.09  & 60.94  & 81.85  & 55.91 & 77.82 & 43.64  & 75.37  & 85.42  & 97.04  & 69.01  & 86.22  & 94.19  & 78.80\\
    LIMN+~\cite{limn}\textcolor{gray}{\scriptsize{ (TPAMI'24)}} & \underline{52.11}  & 75.21  & \underline{57.51}  & \underline{77.92}  & \underline{62.67}  & \underline{82.66}  & \underline{57.43} & \underline{78.60} & 43.33  & 75.41  & 85.81  & 97.21  & 69.28  & 86.43  & 94.26 & 78.79 \\
    IUDC~\cite{iudc}\textcolor{gray}{\scriptsize{ (TOIS'25)}} &35.22& 61.90 &41.86& 63.52& 42.19& 69.23& 39.76& 64.88 & -     & -     & -     & -     & -     & -     & -     & -\\
    QuRe~\cite{QuRe}\textcolor{gray}{\scriptsize{ (ICML'25)}}& 46.80 & 69.81 & 53.53 & 72.87 & 57.47 & 77.77 & 52.60 & 73.48 & \underline{52.22} & \textbf{82.53} & \underline{90.31} & \textbf{98.17} & 78.51 & 91.28 & 96.48 & 84.21 \\
    PAIR~\cite{pair}\scriptsize{\textcolor{gray}{   (ICASSP'25)}} & 46.78  & 70.93  & 52.60  & 73.80  & 58.91  & 78.81  & 52.76  & 74.51 & 46.36 & 78.43 & 87.86 & 97.90 & 74.63 & 89.64 & 95.61 & 81.49\\
    MEDIAN~\cite{median}\scriptsize{\textcolor{gray}{ (ICASSP'25)}} & 46.90  & 70.30  & 52.65  & 73.96  & 57.62  & 78.63  & 52.39  & 74.30 & 45.66 & 78.72 & 87.88 & 97.89 & 75.52 & 89.45 & 95.57 & 81.53\\
    ENCODER~\cite{encoder}\scriptsize{\textcolor{gray}{ (AAAI'25)}}& 51.51  & \underline{76.95}  & 54.86  & 74.93 & 62.01  & 80.88  & 56.13  & 77.59 & 46.10  & 77.98  & 87.16  &  97.64 & 76.92  & 90.41  & 95.95  & 81.74  \\
    \multicolumn{16}{c}{\textbf{\textit{\underline{CVR Models}}}}\\
    CoVR~\cite{covr}\textcolor{gray}{\scriptsize{ (AAAI'24)}} & 44.55  & 69.03  & 48.43  & 67.42  & 52.60  & 74.31  & 48.53 & 70.25 & 49.69  & 78.60  & 86.77  & 94.31  & 75.01  & 88.12  & 93.16  & 80.81\\
    CoVR$\_$Enrich~\cite{covr_enrich}\textcolor{gray}{\scriptsize{ (CVPR'24)}} & 46.12  & 69.52  & 49.61  & 68.88  & 53.79  & 74.74  & 49.84 & 71.05 & 51.03  & -     & 88.93  & 97.53  & 76.51  & -     & 95.76  & -\\
    CoVR-2~\cite{covr-2}\textcolor{gray}{\scriptsize{ (TPAMI'24)}} & 46.53  & 69.60  & 51.23  & 70.64  & 52.14  & 73.27  & 49.97 &71.17 & 50.43 & 81.08 & 88.89 & 98.05 & 76.75 & 90.34 & 95.78 & 83.95 \\
    \hline
           \rowcolor[rgb]{0.949, 0.949, 1.0}
    \textbf{IMAGINE (Ours)}  & \textbf{52.39} & \textbf{77.04} & \textbf{61.61} & \textbf{80.31} & \textbf{63.27} & \textbf{82.99} & \textbf{59.09} & \textbf{80.11} & \textbf{52.25} & \underline{82.39} & \textbf{90.34} & \underline{98.12} & \underline{80.14} & \textbf{92.43} & \textbf{96.92} & \textbf{84.66} \\
    \hline
    \hline
    \end{tabular}%
    }
  \label{tab:cir}%
\end{table*}%

\section{Experiments}\label{sec:experiments}
\subsection{Experimental settings}\label{sec:experiment setting}

\noindent \textbf{Dataset}: We evaluate IMAGINE on three datasets: WebVid-CoVR for the CVR task, and FashionIQ and CIRR for the CIR task. Following standard evaluation criteria, we report R@$k$ ($k$ = 1, 5, 10, 50) and their averages for WebVid-CoVR. For FashionIQ, we provide average values for R@$10$ and R@$50$. For CIRR, we summarize R@$k$ ($k$ = 1, 5, 10, 50) and R$_{sub}$@$k$ ($k$ = 1, 2, 3), providing the average of all seven metrics.

\noindent \textbf{Implementation Details}: Following the previous work~\cite{covr-2,MELT}, we use pre-trained BLIP-2~\cite{blip-2} to extract features. IMAGINE is optimized using the AdamW optimizer with an initial learning rate of 1e-5 on the WebVid-CoVR and FashionIQ datasets, whereas a learning rate of 2e-5 is adopted for CIRR. The hidden dimension $D$ is set to 256. The iteration variable $k$ denotes the global training step, which is equivalent to the epoch count. The top-$T$ selection is set to 10, the modulation coefficient $\rho$ in Equation (\ref{rhomod}) is set to 0.9, and the trade-off hyperparameter $\kappa$ in Equation (\ref{optimization}) is set to 0.5. Model training is performed on a single NVIDIA V100 GPU equipped with 32GB of memory.

\subsection{Perfermance Comparison}
\noindent \textbf{On the CVR task.} As shown in Table~\ref{tab:cvr}, we compare IMAGINE with pre-trained and CVR models. We have obtained the following observations: 1) The results show that IMAGINE outperformed all baselines on the WebVid-CoVR dataset. Specifically, IMAGINE improved R@$1$ by 5.64\% on WebVid-CoVR. These improvements demonstrate that by leveraging semantically related imagery space and assisting feature alignment in the concrete space, IMAGINE enhances multimodal query understanding. 2) Furthermore, IMAGINE surpassed CoVR-Enrich on R@$1$ and R@$5$ in WebVid-CoVR, along with additional improvements in other metrics. Unlike CoVR-Enrich, which relies on external models for textual descriptions and static feature fusion, IMAGINE constructs more accurate query representations by capturing relevant semantics in imagery space, thus mitigating interference from external knowledge hallucinations.

\noindent \textbf{On the CIR task.} To validate IMAGINE's generalization on CIR tasks, we evaluated it on CIR datasets, comparing it with CIR and CVR models (Table \ref{tab:cir}), leading to the following observations: 1) IMAGINE outperformed all models on both datasets. Specifically, it improved the average R@$10$ by 2.89\% on FashionIQ and the Avg metric by 0.26\% on CIRR, demonstrating strong cross-domain generalization. 2) IMAGINE outperformed CIRR more significantly on FashionIQ, likely due to the fashion dataset's reliance on imagery and textual descriptions. While other models may overlook related semantics, leading to irrelevant retrievals, IMAGINE leverages the complementary advantages of the dual space for better results.

\noindent \textbf{Efficiency Comparison.} To validate the efficiency and scalability of IMAGINE, we compared it against the SOTA baseline CoVR-2 on a single NVIDIA V100 GPU. The results demonstrate that IMAGINE achieves a superior balance between performance and efficiency. With a negligible increase in model parameters of 0.95M (approximately 0.08\%), the per-sample inference time of 0.0660s remains comparable to the baseline, ensuring imperceptible inference latency. Although the dynamic construction of the schema-imagery space, which involves iterative updates of inter-modality prototypes and imagery attention calculations, led to a slight increase in training time and memory overhead (approximately 3.0\% and 8.8\% respectively), this marginal cost yields a significant 2.80\% performance gain on WebVid-Avg.

\begin{table}[h]
\centering
\caption{Efficiency Comparison on WebVid-CoVR Dataset.}
\label{tab:efficiency}
\resizebox{\linewidth}{!}{
\begin{tabular}{lccccc}
\hline
\hline
\textbf{Methods} & \textbf{Params (M)} & \textbf{Test (s/sample)} & \textbf{Memory} & \textbf{Train (s/iteration)} & \textbf{WebVid-Avg} \\ \hline
CoVR-2 & 1173.19M & 0.0653 & 18541M & 9.612 & 83.30\\
\textbf{IMAGINE} & 1174.14M & 0.0660 & 20173M & 9.901 & 85.63 \\ 
\hline
\hline
\end{tabular}
}
\end{table}

\subsection{Ablation Study}\label{sec:ablation studies}

To comprehensively validate the effectiveness and synergy of the core components within the IMAGINE framework, we conducted extensive ablation studies on the WebVid-CoVR, FashionIQ, and CIRR datasets. The experimental design was divided into four primary groups to investigate the individual contributions and internal design efficacy of \textit{Schema Imagery Construction (SIC)}, \textit{Imagery-guided Multimodal Composition (IMC)}, and \textit{Dual Space Alignment (DSA)}. Overall, the full model achieved optimal performance across all evaluation metrics, whereas the exclusion of any critical module resulted in decreased retrieval accuracy. These results compellingly demonstrate that explicitly modeling the imagery space and ensuring its deep alignment and fusion with the concrete space are pivotal for CVR performance in complex scenarios. The subsequent sections provide a comprehensive in-depth analysis of specific impacts through various module variants.

\textbf{Loss.} Based on the ablation studies presented in Table \ref{tab:ablation-loss}, we validated the effectiveness of the core modules within the IMAGINE framework. The results indicate that the full model achieves optimal performance across all metrics, confirming the indispensability of each component. Specifically, removing the concrete space classification loss ($w/o\ \mathcal{L}_{itc}$) caused a precipitous performance drop, with WebVid-Avg falling to 47.00\%, confirming that explicit object correspondence serves as the discriminative foundation for CVR tasks. Furthermore, removing \textit{Schema Imagery Construction} ($w/o\ SIC$) significantly reduced WebVid-Avg to 82.86\%, demonstrating the necessity of explicitly modeling the imagery space to complement implicit semantics. Notably, the performance loss resulting from removing only the imagery alignment constraint within \textit{Dual Space Alignment} ($w/o\ \mathcal{L}_{schema}$), which dropped to 82.77\%, was slightly higher than that of completely removing the SIC module. This result strongly suggests that without explicit guided alignment from the imagery space, constructed imagery prototypes easily deviate from target semantics and introduce detrimental interfering noise, thereby substantially impairing retrieval accuracy and underscoring the critical role of joint dual-space alignment.

\begin{table}[h]
  \centering
  \caption{Ablation study of different components on FashionIQ, WebVid-CoVR, and CIRR datasets.}
  \resizebox{\linewidth}{!}{
    \begin{tabular}{ccc|cc|c|c}
    \hline
    \hline
    \multirow{2}{*}{$\mathcal{L}_{itc}$} & \multirow{2}{*}{$\mathcal{L}_{schema}$} & \multirow{2}{*}{w/o SIC} & \multicolumn{2}{c|}{FIQ-Avg.} & WebVid & CIRR \\
\cline{4-7}          &       &       & Avg.R@10 & Avg.R@50 & Avg.  & Avg. \\
    \hline
    \checkmark     &       &       & 54.98  & 75.64  & 80.47  & 81.21  \\
          & \checkmark     &       & 13.32  & 28.53  & 35.77  & 35.93  \\
    \hline
          & \checkmark     & \checkmark     & 18.78  & 34.37  & 47.00  & 39.19  \\
    \checkmark     &       & \checkmark     & 56.80  & 77.59  & 82.77  & 82.04  \\
    \checkmark     & \checkmark     &       & 57.09  & 78.03  & 82.86  & 82.21  \\
    \hline
    \rowcolor[rgb]{0.949, 0.949, 1.0}
    \checkmark     & \checkmark     & \checkmark     & \textbf{59.09} & \textbf{80.11} & \textbf{85.63} & \textbf{84.66} \\
    \hline
    \hline
    \end{tabular}%
  \label{tab:ablation-loss}%
  }
\end{table}%

\begin{figure}[h]
  \begin{center}
  \includegraphics[width=\linewidth]{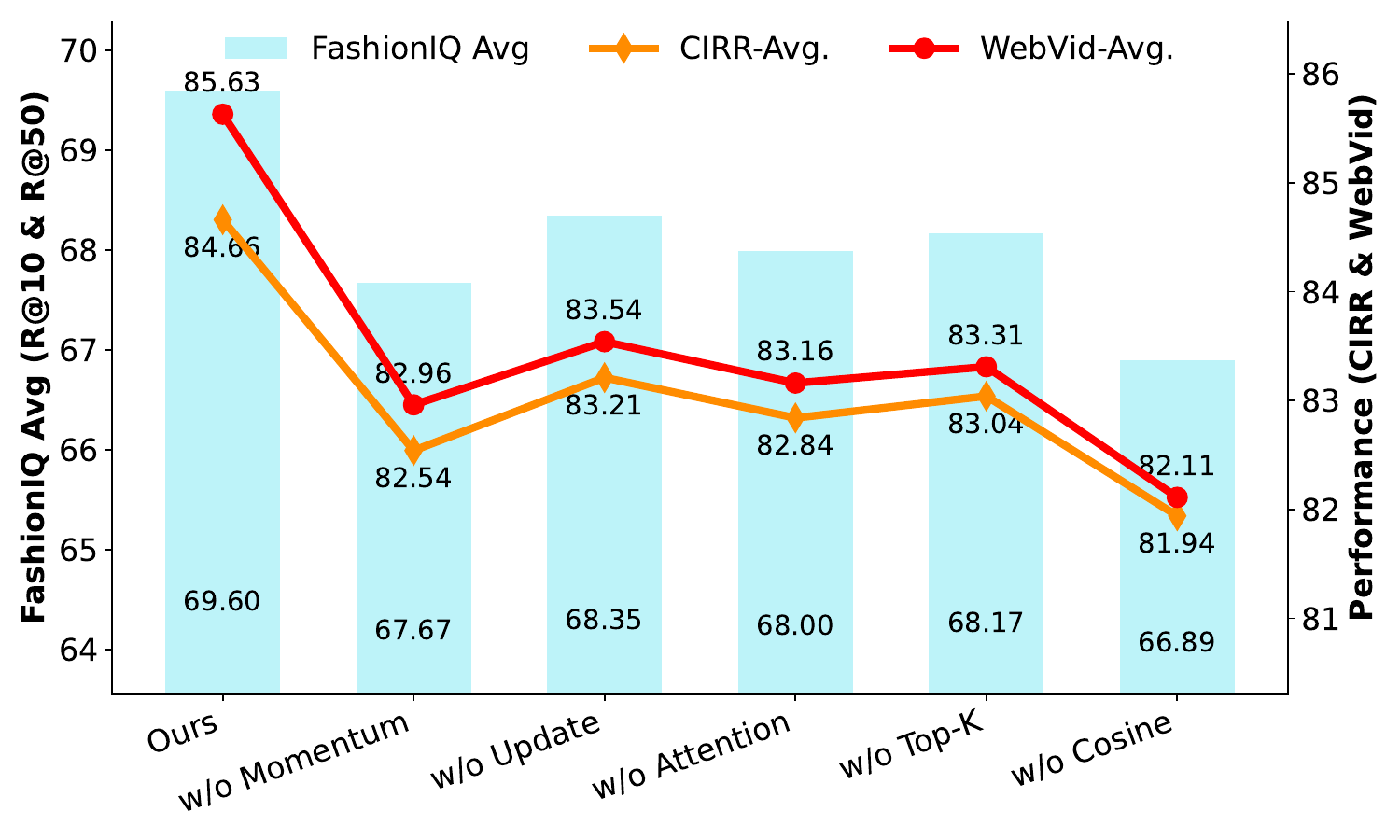}
  \end{center}
  \caption{Ablation study on the internal designs of the Schema Imagery Construction (SIC) module.}
  \label{fig:ablation-sic}
\end{figure}

\textbf{SIC.} Figure \ref{fig:ablation-sic} validates that the synergy among components within the SIC module is critical for high-quality imagery generation. First, employing a random selection strategy ($w/o\ cosine$) caused the WebVid average score to plummet to $82.11\%$. This confirms that maintaining strict semantic relevance is fundamental to constructing the imagery space, whereas indiscriminately introducing noise from irrelevant prototypes severely undermines imagery construction. Second, discarding the batch momentum update ($w/o\ Momentum$) resulted in a significant performance decline (FIQ dropped to $67.67\%$), performing worse than static prototypes ($w/o\ Update$, $68.35\%$). This demonstrates that aggressive updates introduce statistical noise within batches, whereas the momentum update strategy is essential for maintaining the evolutionary stability of the prototype library. Third, replacing attention aggregation with average aggregation ($w/o\ Attention$) reduced the WebVid score to $83.16\%$. This establishes that the model must utilize adaptive attention weights to distinguish the importance of different prototypes relative to the current query to precisely focus on core imagery semantics. Finally, removing Top-K filtering ($w/o\ Top-K$) decreased the CIRR average score by $2.7\%$, proving that filtering low-confidence redundant prototypes effectively prevents long-tail semantics from interfering with shared imagery vectors.

\begin{figure}[h]
  \small
  \begin{center}
  \includegraphics[width=0.93\linewidth]{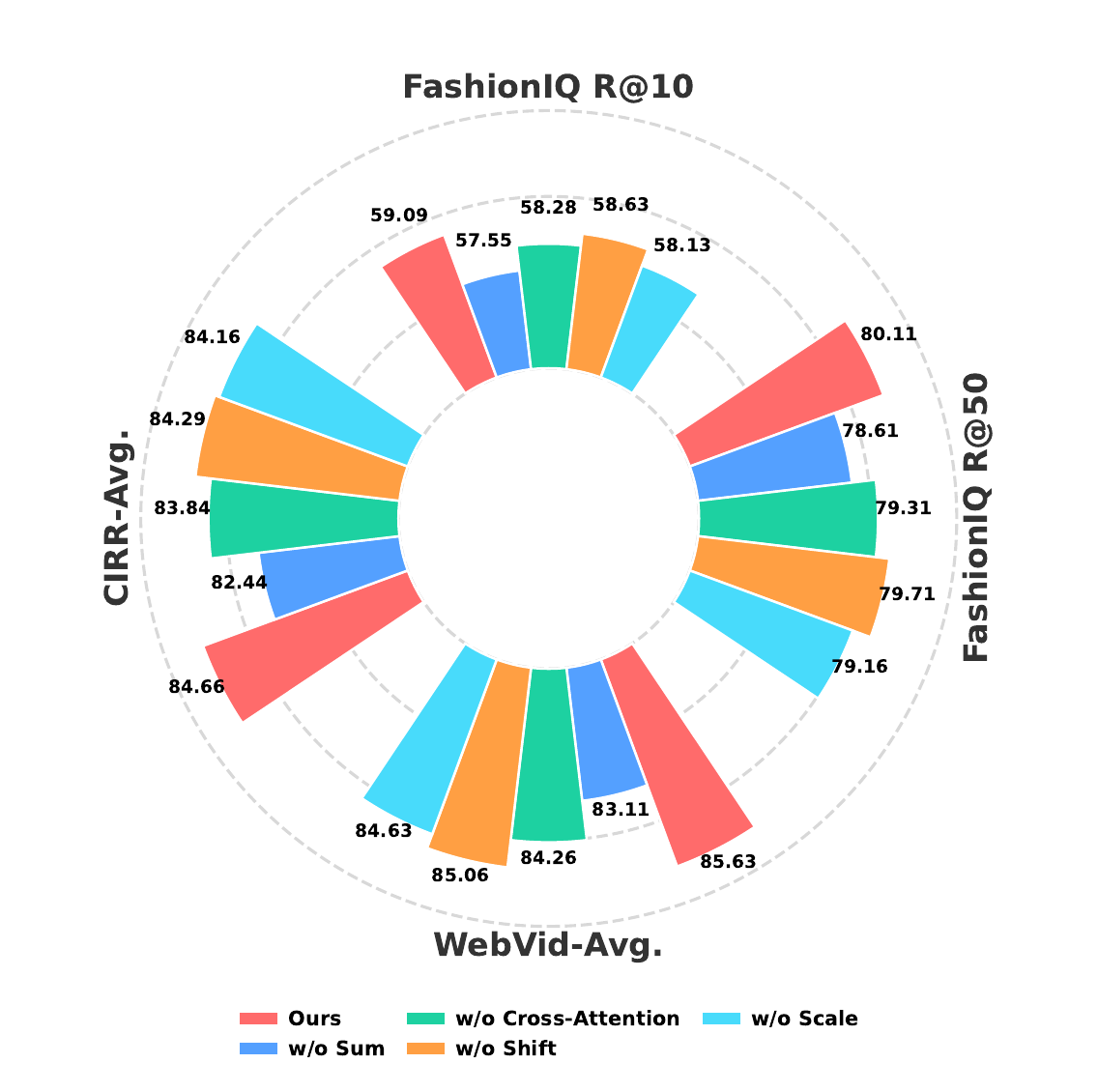}
  \end{center}
  \caption{Impact of different interaction mechanisms and modulation factors in the Imagery-guided Multimodal Composition (IMC) module.}
  \label{fig:ablation-imc}
\end{figure}

\textbf{IMC.} The ablation analysis of the internal components of the IMC module shown in Figure \ref{fig:ablation-imc} further reveals the critical role of the refined modulation mechanism in cross-space semantic fusion. First, degrading the modulation mechanism to element-wise addition ($w/o\ Sum$) caused FashionIQ metrics to decline to 57.55\% and 78.61\%, respectively. This confirms that simple feature superposition cannot achieve confidence-based dual-space modulation and fails to adaptively control the integration of imagery information. Second, employing cross-attention ($w/o\ Cross-Attention$) as a substitute for the modulation network reduced CIRR performance to 83.84\%. This indicates that compared to complex attention calculations, the lightweight modulation mechanism more precisely achieves the adaptive injection of implicit semantics without introducing additional noise. Third, the performance loss on WebVid resulting from removing the scaling factor ($w/o\ Scale$) (dropping to 84.63\%) exceeded that of removing the shifting factor ($w/o\ Shift$, 85.06\%). This establishes the dominant role of dynamic scaling in adjusting the dimensional importance of concrete features, whereas the shifting operation performs an auxiliary refinement function.

\begin{figure}[h]
  \begin{center}
  \includegraphics[width=\linewidth]{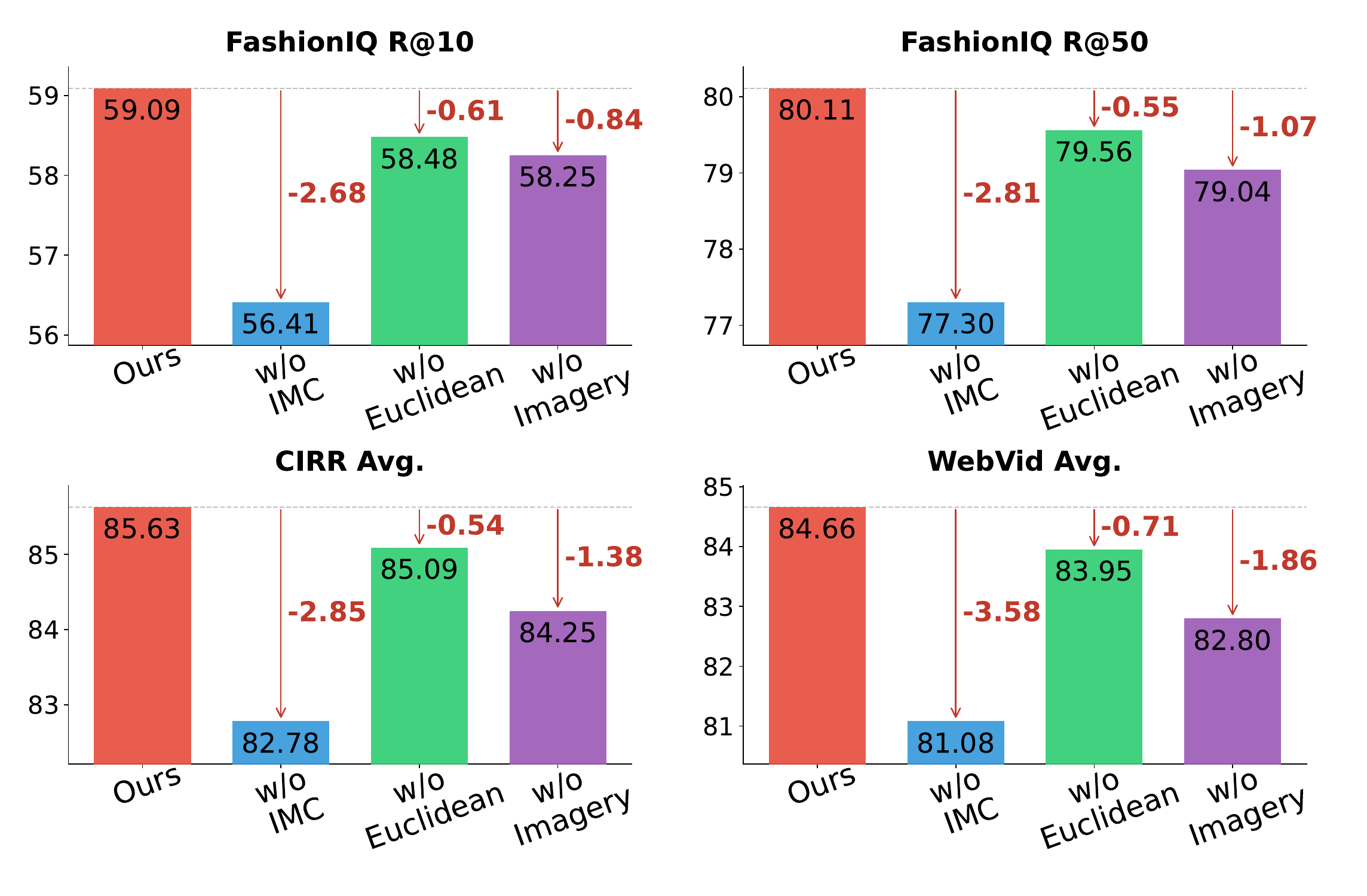}
  \end{center}
  \caption{Ablation study on the Dual Space Alignment (DSA) module and the impact of the IMC module.}
  \label{fig:ablation-dsa}
\end{figure}

\textbf{DSA.}Figure \ref{fig:ablation-dsa} validates the effectiveness of the critical processing applied to the imagery space within IMAGINE. Specifically, removing the \textit{Imagery-guided Multimodal Composition (IMC)} module ($w/o\ IMC$) led to the most drastic performance degradation across all tasks (e.g., a 3.58\% drop in WebVid-Avg). This firmly establishes the core status of the module and demonstrates that injecting implicit imagery semantics into the concrete space via confidence-based modulation is crucial for resolving semantic-visual misalignment. Second, concerning the internal mechanisms of the \textit{Dual Space Alignment (DSA)} module, altering the imagery alignment target ($w/o\ Imagery$) resulted in the second most significant decline (e.g., a 1.38\% drop in CIRR). This confirms that dual space alignment must adhere to a consistency principle; forcing the alignment of imagery vectors to heterogeneous concrete targets undermines the independent representation capability of the imagery space. Finally, substituting the metric with Euclidean distance ($w/o\ Euclidean$) in the DSA module caused a 0.61\% drop on FashionIQ. This indicates that within the abstract imagery semantic space, direction-based cosine similarity is more effective at capturing semantic associations than absolute distance metrics.

\subsection{Parameter Sensitivity}

\begin{figure}[h]
  \begin{center}
  \includegraphics[width=\linewidth]{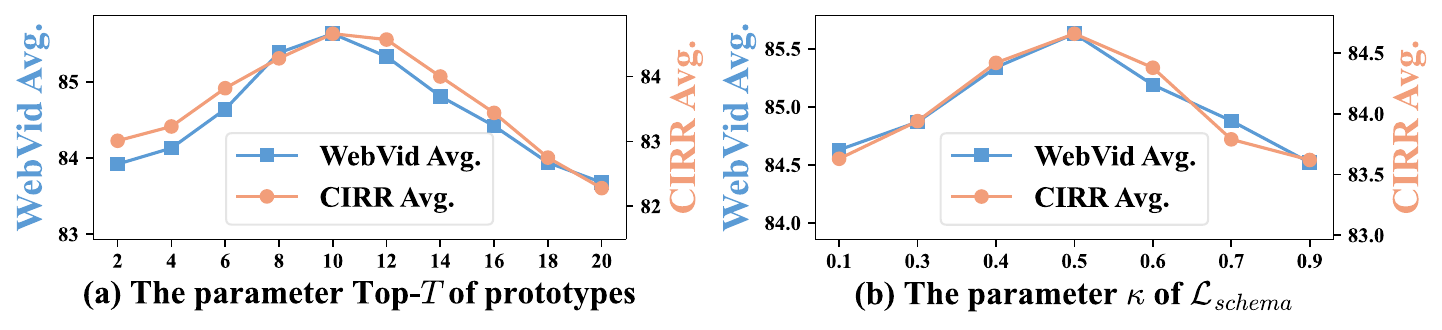}
  \end{center}
  \caption{Parameter sensitivity experiments of IMAGINE on the WebVid-CoVR and CIRR datasets. The left figure shows the variation in the number of Top-T prototypes, and the right figure displays the hyperparameters of the loss function $\mathcal{L}_{schema}$.}
  \label{fig:Parameter}
\end{figure}

We conducted a comprehensive parameter sensitivity analysis on the number of selected imagery prototypes Top-$T$, the schema loss weight $\kappa$, and the modulation coefficient $\rho$ for the prototype library update. As illustrated in Figure \ref{fig:Parameter} and Table \ref{tab:rho}, all three hyperparameters exhibit a consistent trend where performance initially increases to a peak before declining.

First, regarding the number of selected imagery prototypes Top-$T$, the model achieves optimal performance at $T=10$ as shown in Figure \ref{fig:Parameter}(a). An insufficient $T$ value unduly restricts the diversity of retrieved imagery prototypes, preventing the model from capturing sufficiently comprehensive and rich implicit semantics. Conversely, an excessively large $T$ value (e.g., $T>14$) introduces numerous low-relevance redundant prototypes. These detrimental noises dilute the expression of core imagery semantics, thereby impairing the construction quality of the imagery space.

Second, concerning the loss weight $\kappa$, the best results occur at $0.5$ as shown in Figure \ref{fig:Parameter}(b). A lower $\kappa$ value (e.g., $0.1$) weakens the guiding effect of imagery space alignment, making it difficult for the model to utilize latent imagery semantics to bridge the semantic-visual misalignment between the reference video and the target video. In contrast, an excessively large $\kappa$ value causes the optimization process to overemphasize implicit imagery consistency, which to some extent overshadows the role of the primary classification loss $\mathcal{L}_{itc}$ in the concrete space, affecting the discriminative ability of explicit visual features.

Finally, as indicated in Table \ref{tab:rho}, the model reaches peak performance across all metrics when $\rho=0.9$. Smaller $\rho$ values (e.g., $0.4-0.8$) imply that features from the new batch carry excessive weight in the prototype update, causing the prototype library to overfit the statistical distribution of the current batch and introduce unnecessary noise, thereby destabilizing the imagery manifold. Notably, performance drops significantly when $\rho=1.0$. This result strongly validates that differences exist in imagery features corresponding to different samples; therefore, the prototype library must be continuously iteratively updated during training to accurately capture evolving imagery semantics. In summary, based on the experimental results, we set Top-$T$ to 10, $\kappa$ to 0.5, and $\rho$ to 0.9.

\begin{table}[htbp]
  \centering
  \small
  \tabcolsep=10pt
  \caption{Performance result of IMAGINE under different modulation coefficients $\rho$.}
  \resizebox{0.9\linewidth}{!}{
    \begin{tabular}{c|cc|c|c}
    \hline
    \hline
    \multirow{2}{*}{$\rho$} & \multicolumn{2}{c|}{FIQ-Avg.} & WebVid & CIRR \\
\cline{2-5}          & Avg.R@10 & Avg.R@50 & Avg.  & Avg. \\
    \hline
    \hline
    0.4   & 57.18  & 77.86  & 83.73  & 83.03  \\
    0.5   & 57.75  & 78.51  & 84.23  & 83.76  \\
    0.6   & 58.38  & 79.28  & 84.94  & 84.01  \\
    0.8   & 58.91  & 79.91  & 85.25  & 84.49  \\
    \rowcolor[rgb]{0.949, 0.949, 1.0}
    0.9   & 59.09  & 80.11  & 85.63  & 84.66  \\
    1     & 57.31  & 77.98  & 83.72  & 83.21  \\
    \hline
    \hline
    \end{tabular}%
  \label{tab:rho}%
  }
\end{table}%

\subsection{Qualitative Results}

\begin{figure}[h]
  \begin{center}
  \includegraphics[width=\linewidth]{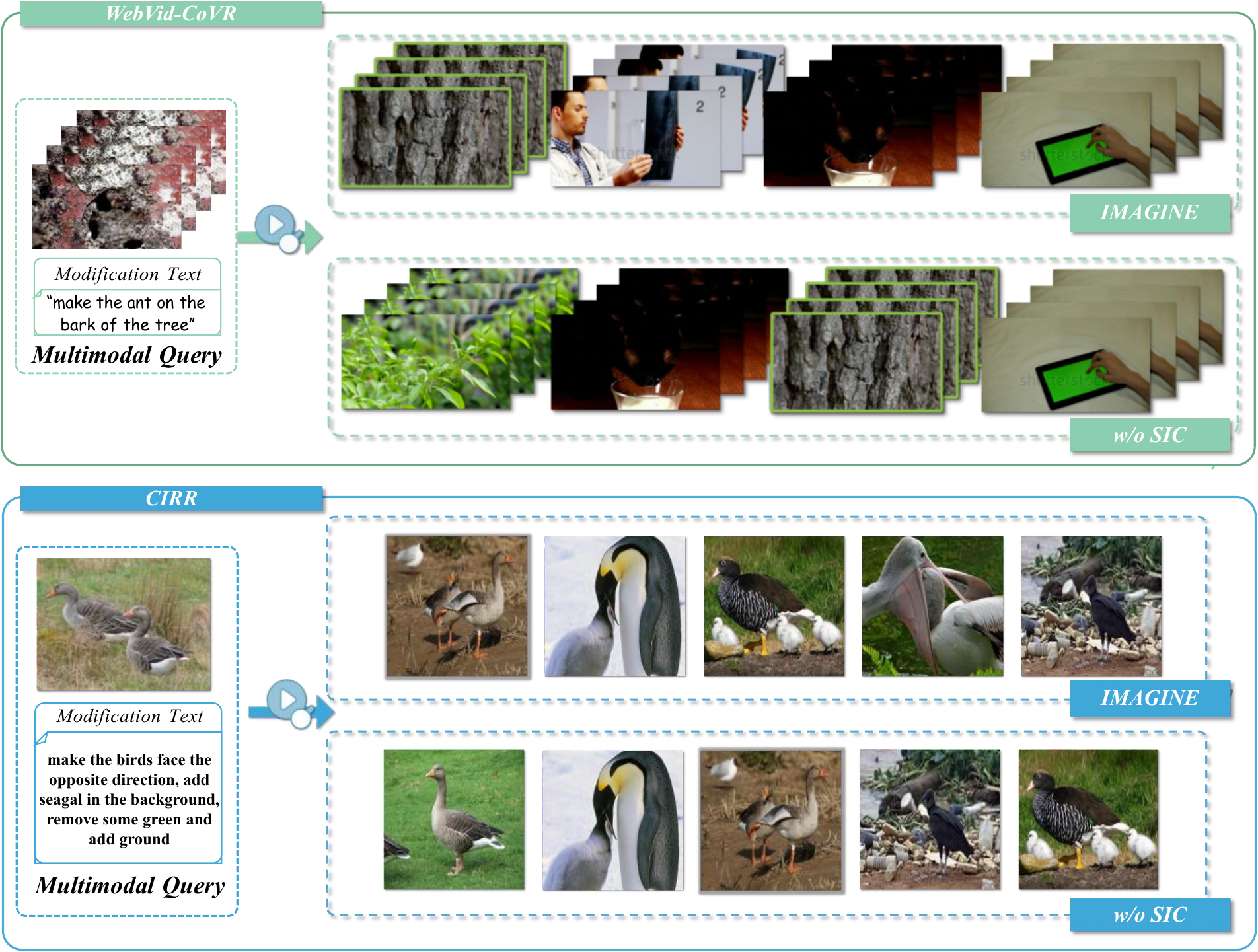}
  \end{center}
  \caption{Qualitative comparison on WebVid-CoVR and CIRR datasets. We compare the retrieval results of IMAGINE with the baseline without the SIC module (w/o SIC). The green boxes indicate the ground-truth target videos or images.}
  \label{fig:case}
\end{figure}

\textbf{Case Study.}The visualization results in Figure \ref{fig:case} illustrate the significant advantages of IMAGINE in addressing semantic-visual misalignment. Compared with the baseline model (w/o SIC) constrained by explicit visual features, IMAGINE exhibits superior cross-space reasoning capabilities. For instance, in the WebVid-CoVR case, given the instruction ``move ants to bark'', the baseline model mistakenly retrieved generic vegetation scenes due to the lack of a visual reference for ``bark''. In contrast, IMAGINE utilizes the \textit{Schema Imagery Construction (SIC)} module to actively ``imagine'' and retrieve texture-related implicit semantic prototypes within the imagery space, successfully guiding the model to locate the target video containing rough bark. Similarly, in the CIRR case involving the complex requirements of ``add seagulls'' and ``change background to ground'', IMAGINE was not misled by the grass in the reference image; instead, it leveraged supplementary information from the imagery space to accurately reconstruct the target scene, capturing both specific actions and the wasteland atmosphere. This fully demonstrates that compensating for the absence of concrete information via the imagery space is key to enhancing retrieval accuracy for complex retrieval intentions.

\begin{figure}[h]
  \begin{center}
\includegraphics[width=\linewidth]{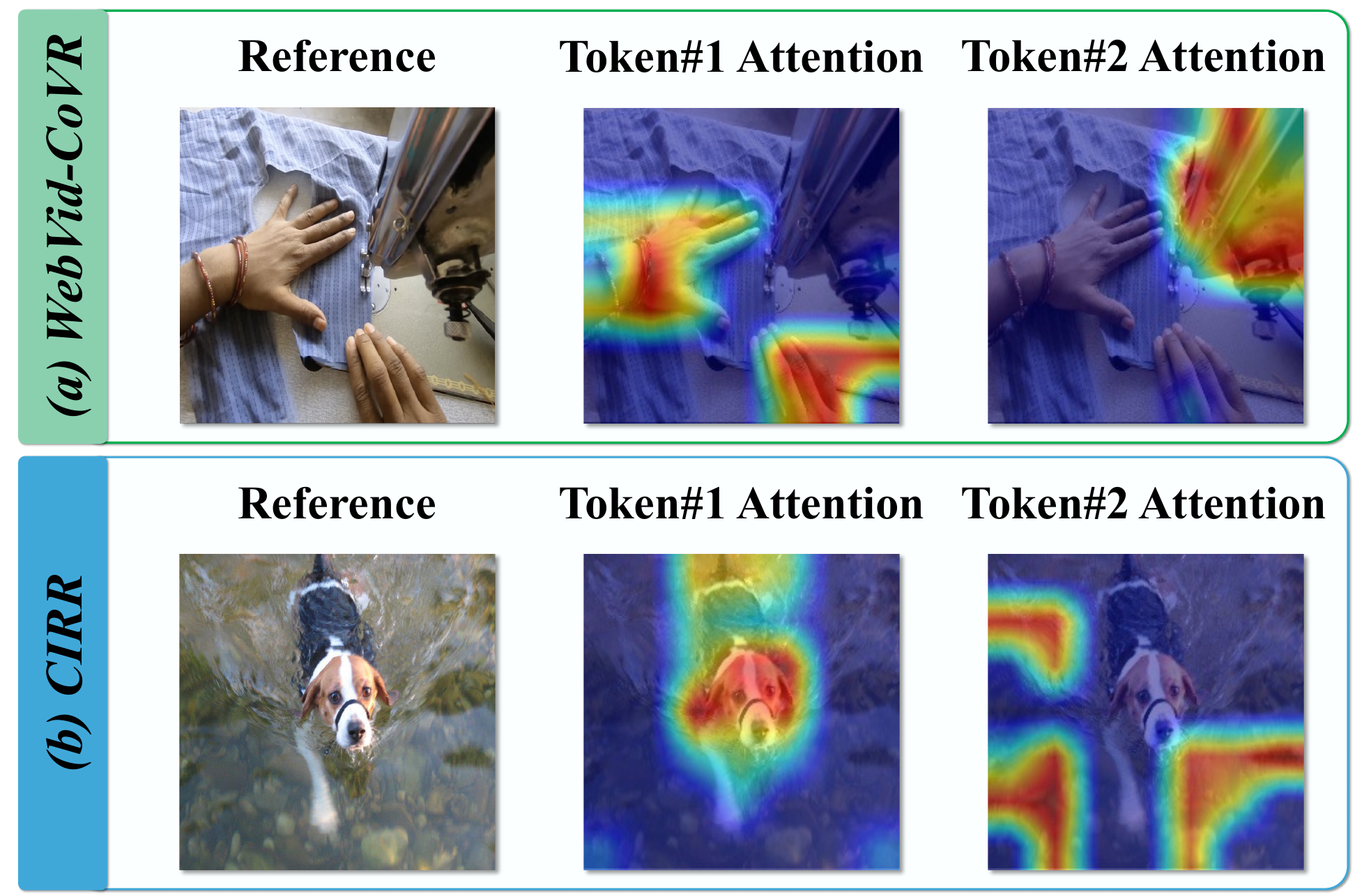}
  \end{center}
  \caption{Visualization of the attention maps from the SIC module on WebVid-CoVR and CIRR datasets.}
  \label{fig:attention_vis}
\end{figure}

\noindent\textbf{Attention Visualization.}As shown in Fig.~\ref{fig:attention_vis}, we visualize imagery token attention maps to verify the SIC module's ability to capture diverse semantic clues. The core motivation of IMAGINE is to construct a robust imagery space that complements concrete features. Visualization results reveal a disentangled attention mechanism: in WebVid-CoVR Fig.~\ref{fig:attention_vis} part (a), different tokens focus on complementary visual concepts, such as dynamic actions versus static tools. This indicates the generated imagery is a composite of fine-grained semantic details rather than a monolithic representation. Furthermore, the model demonstrates strong context-awareness; in the CIRR case, Fig.~\ref{fig:attention_vis} part (b), tokens effectively separate foreground subjects from background contexts. These distinct patterns demonstrate that the SIC module adaptively extracts multi-view semantic information instead of merely replicating global features. This distinction enables the IMC module to reason effectively about complex modification requests.

\section{Conclusion}

\noindent In this work, we identify a key limitation of existing CVR models, namely the common assumption that the modified object directly appears in the reference video. This assumption overlooks the possibility that some modified objects may appear in the video in the form of semantically related imagery. To address this issue, we propose a new adaptive scheme, IMAGINE, to solve the CVR task. IMAGINE can adaptively construct the imagery space and simultaneously guide the feature alignment in the concrete space to achieve better retrieval performance. Our IMAGINE model achieves optimal results across all metrics on three widely used benchmark datasets, covering both CVR and CIR tasks.

\begin{acks}
   This work was supported in partby the National Natural Science Foundation of China, No.:62576195, and No.:62276155; in part by the Key R\&D Program of Shandong Province (Major scientific and technological innovation projects), China, No.: 2025CXGC020101; in part by the China National University Student Innovation \& Entrepreneurship Development Program, No.:2025283 
\end{acks}

\bibliographystyle{ACM-Reference-Format}
\bibliography{main}

\end{document}